\documentclass{article}

 \usepackage[preprint]{neurips_2026}

\usepackage{listings}
\lstdefinelanguage{JSON}{
  basicstyle=\ttfamily\small,
  breaklines=true,
  showstringspaces=false,
  columns=fullflexible,
  frame=single,
  numbers=left,
  numberstyle=\tiny\color{gray},
  keywordstyle=\color{blue},
  stringstyle=\color{red},
  commentstyle=\color{green}
}
\usepackage{tocloft}
\usepackage{titletoc}
\usepackage{etoolbox}
\usepackage{titlesec}

\usepackage{amsmath,amsfonts,bm}

\def\eqref#1{equation~\ref{#1}}
\def\1{\bm{1}}

\DeclareMathAlphabet{\mathsfit}{\encodingdefault}{\sfdefault}{m}{sl}
\SetMathAlphabet{\mathsfit}{bold}{\encodingdefault}{\sfdefault}{bx}{n}

\usepackage{url}
\usepackage[final,nopatch=footnote]{microtype}
\usepackage[export]{adjustbox}
\usepackage{graphicx}
\usepackage{subfig}
\usepackage{tikz}
\usepackage{bbm}
\usepackage{booktabs} % for professional tables
\usepackage{enumitem}
\usepackage{xargs}
\usepackage{algorithm}
\usepackage[noend]{algpseudocode}
\usepackage{eqparbox}
\usepackage{multirow}
\usepackage{makecell}
\usepackage{amsmath,amssymb,mathtools}
\usepackage{bigints}
\usepackage{dsfont}
\usepackage{wrapfig}
\usepackage{natbib}
\usepackage{varwidth}% http://ctan.org/pkg/varwidth
\usepackage[colorlinks, linkcolor=myblue, citecolor=gray]{hyperref}
\usepackage{float}
\usepackage{nicefrac}       % compact symbols for 1/2, etc.
\usepackage{xurl}

\usepackage[most]{tcolorbox}
\usepackage{pifont}
\usepackage[table]{xcolor}

\usepackage{stackengine}
\usepackage{relsize,exscale}
\usepackage{mathrsfs}

\title{Adaptive Nucleus Truncation for Long-Form Reasoning}

\author{%
  Ousmane Amadou~Dia%\thanks{Use footnote for providing further information
  \\
  \texttt{ousamdia@gmail.com} \\
}

\definecolor{mydarkblue}{rgb}{0,0.08,0.45}
\definecolor{mydarkred}{rgb}{0.6,0,0}
\definecolor{myblue}{HTML}{268BD2}
\definecolor{mygreen}{HTML}{658354}

\definecolor{plot_blue}{HTML}{8DC3E5}
\definecolor{plot_green}{HTML}{A4D8A7}
\definecolor{plot_yellow}{HTML}{EFDAA4}
\definecolor{plot_red}{HTML}{DEABB9}

\tcbset{
  mm/prompt/base/.style={
    enhanced, breakable,
    colback=gray!5!white,
    colframe=gray!75!black,
    boxrule=0.6pt, arc=0.8mm,
    left=1.2mm, right=1.2mm, top=1.2mm, bottom=1.2mm,
    before skip=6pt, after skip=6pt,
    fonttitle=\large\bfseries, coltitle=white,
    title={},               
    before upper=\setlength{\parskip}{2pt}\footnotesize
  },
  mm/prompt/light/.style={
    enhanced, breakable,
    colback=gray!4!white,
    colframe=gray!60!black,
    boxrule=0.6pt, arc=0.8mm,
    left=1.2mm, right=1.2mm, top=1.2mm, bottom=1.2mm,
    before skip=6pt, after skip=6pt,
    fonttitle=\bfseries, coltitle=black,
    title={},
    before upper=\setlength{\parskip}{2pt}\small
  }
}

\newtcolorbox{PromptBox}[1][]{mm/prompt/base,#1}
\newtcolorbox{PromptBoxLight}[1][]{mm/prompt/light,#1}

\begin{document}

\maketitle

\begin{abstract}
Sampling plays an important role in long-form language-model reasoning. Over thousands of decoding steps, small changes in the candidate token set can compound into different reasoning trajectories, stability profiles, and final answers. Existing truncation methods such as top-$p$, min-$p$, and fixed top-$n\sigma$ sampling improve over unrestricted sampling, but they rely on fixed thresholds that cannot adapt to changes in entropy, task difficulty, training stage, or generation budget. We introduce Adaptive Nucleus Truncation Sampling (ANTS), which extends top-\(n\sigma\) sampling from a fixed decoding rule into an adaptive rollout-control mechanism for long-form generation. ANTS selects standardized neighborhoods around the maximum logit before temperature scaling, adapts the truncation width using an entropy-conditioned controller, and retains a no-truncation fallback arm to stabilize training when truncation becomes unsafe. On a 33B-total / 4B-active sparse Mixture-of-Experts reasoning model, ANTS improves average performance over percentage-based benchmarks by +1.9, +3.8, and +5.2 points at 8K, 16K, and 32K generation budgets, respectively. The strongest gains appear on instruction following and mathematical reasoning, with IFBench improving by more than 10 points at 32K and AIME 2025 improving by 7 points. Code generation reveals an important budget interaction. On Codeforces, ANTS trails the baseline at 8K, but reverses this gap and substantially improves ELO at 16K and 32K. These results suggest that sampler design should be treated not just as a decoding hyperparameter, but as part of how we stabilize and scale long-budget reasoning.

\end{abstract}

\section{Introduction}
Large language models (LLMs) are increasingly evaluated in regimes where the generation budget is large enough to express multi-step reasoning, self-refinement, tool plans, and executable code~\citep{wei2022chain, madaan2023selfrefine, yao2023react, chen2021codex, li2022alphacode}. In these settings, decoding is not a minor post-processing detail. The sampler controls which reasoning paths are reachable, how much stochasticity is exposed to the policy during RL rollouts, and how often a long response drifts into noisy, irrelevant, or unstable continuations. Classical decoding heuristics such as top-$k$, nucleus sampling, and min-$p$ attempt to balance quality and diversity by restricting the candidate token set before sampling~\citep{Holtzman2020The,nguyen2025turningheatminpsampling}. However, probability-thresholding methods such as top-$p$ and min-$p$ operate after softmax, and therefore inherit the distortions introduced by probability normalization and temperature scaling. As a result, the same underlying logit geometry may produce substantially different candidate sets as temperature changes.

\cite{tang-etal-2025-top} recently argued that informative and noisy tokens are more cleanly separated in pre-softmax logit space than in probability space. Their top-$n\sigma$ sampler keeps tokens whose logits fall within a standard-deviation window of the maximum logit, yielding a candidate set that is invariant to positive logit rescaling. This insight provides the geometry we need, but not the policy for how aggressively to use it. A single fixed $n$ cannot simply serve all token positions, tasks, checkpoints, and generation budgets. Some contexts are decisive and should be pruned sharply. Others are genuinely ambiguous and require a wider support. In unstable RL phases, no truncation may be preferable at all.

In this paper, we turn top-\(n\sigma\) from a fixed decoding heuristic into a budget-aware rollout controller. We call the resulting method \emph{Adaptive Nucleus Truncation Sampling} (ANTS). Under ANTS, sampling can be decomposed into three pieces: (i) logit-space support estimation, (ii) context-dependent thresholding, and (iii) stability control. First, ANTS selects the top-logit neighborhood as follows:
\begin{equation}
\mathcal{N}_t(n) = \Big\{v \in \mathcal{V}\,\Big\vert\, \ell_{t,v} > \max_{u \in \mathcal{V}} \ell_{t,u} - n \cdot \sigma(\ell_t)\Big\},
\label{eq:ants_nucleus}
\end{equation}
where \(\ell_t\) is the pre-temperature logit vector and \(\sigma(\ell_t)\) is its vocabulary-wise (\(\mathcal{V}\)) standard deviation. Second, it makes the width of this neighborhood entropy-conditioned, \(n_t = n_0 + \gamma\cdot \mathcal{H}(p_t^{(0)})\), where \(p_t^{(0)}=\texttt{softmax}(\ell_t)\) is the unit-temperature distribution. Finally, it learns the truncation strength online with a Thompson-sampling bandit over finite \(\gamma\) values and one explicit fallback arm \(K+1\) with \(\gamma_{K+1}=+\infty\). Selecting the fallback sets \(\mathcal{N}_t=\mathcal{V}\), exactly recovering the baseline sampler.

The fallback arm is central to our method. Without it, training can start well under truncation but later overshoot in entropy, KL, gradient norm, log-ratio statistics, and related diagnostics because every rollout remains subject to some finite pruning pressure. The fallback, $\gamma_{K+1}=+\infty$, gives the controller a learned way to stop truncating before these metrics become unstable. Thus, unlike fixed top-$n\sigma$, ANTS does not assume that truncation is always beneficial for decoding. In fact, it learns when truncation helps and when the safest action is to return to the untruncated rollout distribution.

Our main empirical finding is that ANTS improves with generation budget. This effect is clearest on IFBench and AIME 2025, where longer generations create more opportunities for verbose, irrelevant, or unstable continuations to accumulate. ANTS appears to preserve enough diversity for reasoning while suppressing tails that lead to unnecessary continuation. Codeforces provides the main caveat at 8K. The issue is not simply that truncation shortens completions. In fact, ANTS generates more solution tokens in this regime, yet trails the baseline in ELO. A more plausible explanation is that, under a tight budget, the induced distribution is less favorable for reaching correct complete programs. At 16K and 32K generation lengths, where the feasible program space is less constrained, the effect reverses into large ELO gains. This suggests that truncation should be interpreted jointly with generation budget, task type, and training state rather than as a fixed decoding hyperparameter.

% \textbf{Contributions.} We make the following contributions. We formulate ANTS as an entropy-conditioned, temperature-invariant logit-space sampler for RL rollouts. We introduce a simple bandit controller that adapts truncation strength online and retains a no-truncation fallback. We provide a context-length scaling evaluation across 8K, 16K, and 32K budgets, showing that gains generally grow with the generation budget and are especially strong for instruction following. We characterize an important failure mode on short-budget code generation to motivate more budget-aware truncation policies.
\textbf{Contributions.} We formulate ANTS as an adaptive logit-space sampler with temperature-invariant support selection for RL rollouts. First, we introduce an entropy-driven Thompson controller that adapts truncation strength online and retains a no-truncation fallback arm. Second, we provide a generation-budget scaling evaluation across 8K, 16K, and 32K budgets, showing that gains generally grow with the generation budget and are especially strong for instruction following. Third, we characterize a short-budget code-generation caveat that motivates budget-aware truncation policies.
% \begin{enumerate}[leftmargin=1.5em,itemsep=0.2em]
%     \item We formulate ANTS, an entropy-conditioned, temperature-invariant logit-space sampler for evaluation and RL rollouts.
%     \item We introduce a simple bandit controller that adapts truncation strength online and retains a no-truncation fallback.
%     \item We provide a context-length scaling evaluation across 8K, 16K, and 32K budgets, showing that gains generally grow with the generation budget and are especially strong for instruction following.
%     \item We characterize an important failure mode on short-budget code generation, motivating budget-aware truncation policies.
% \end{enumerate}

\section{Background and Motivation}

\textbf{Probability-space truncation.} Nucleus sampling selects the smallest prefix of the probability-sorted vocabulary whose cumulative mass exceeds a threshold \(p\) \citep{Holtzman2020The}. Min-\(p\) instead uses the top token's probability as a confidence-scaled threshold \citep{nguyen2025turningheatminpsampling}. While these methods are simple and effective, they both operate after softmax. Consequently, their candidate sets depend on the temperature-transformed distribution \(\mathrm{softmax}(\ell_t/\tau)\). For long generations, small differences in candidate-set construction can compound over many steps and destabilize training.

\textbf{Logit-space truncation.} The top-\(n\sigma\) sampling algorithm, introduced by \citep{tang-etal-2025-top}, replaces probability thresholds with standardized distances in logit space. This is useful because the membership of Equation~\ref{eq:ants_nucleus} is invariant to positive rescaling of logits. More formally, for any \(\omega>0\),
\begin{equation}
\ell_{t,v} > \max_u \ell_{t,u} - n\sigma(\ell_t)
\quad \Longleftrightarrow \quad
\omega\ell_{t,v} > \max_u \omega\ell_{t,u} - n\sigma(\omega\ell_t).
\end{equation}
Temperature scaling corresponds to setting \(\omega=\nicefrac{1}{\tau}\) before softmax. Thus, if the nucleus set \(\mathcal{N}_t\) is selected before the logits are scaled by the temperature, the set of candidate tokens is temperature-invariant. More intuitively, the temperature only reshapes the probabilities inside the selected set.

\textbf{Entropy conditioning.} A fixed top-\(n_t\) threshold is insufficient across token positions. Some contexts are decisive, with one or a few plausible continuations. Others may, however, be genuinely ambiguous. Thus, the same fixed \(n\) cannot simultaneously serve mathematical derivations, instruction-following constraints, and code completions under different generation budgets. ANTS therefore adapts \(n_t\) with a base entropy signal. Since Equation~\ref{eq:ants_nucleus} keeps tokens within \(n_t\) standard deviations of the maximum logit, increasing \(n_t\) makes the filter less aggressive. High-entropy contexts therefore retain more alternatives, while low-entropy contexts are pruned more sharply. This turns top-\(n_t\) from a fixed decoding heuristic into a context-dependent support estimator, motivating the entropy conditioning.

\textbf{Online controller.} The optimal truncation strength depends on task, generation budget, and training checkpoint. Rather than hard-coding a single \(\gamma\), we cast truncation selection as a Thompson-style controller over a finite arm set. Because the candidate distributions for all arms can be evaluated on the same sampled logits, each arm receives an intrinsic entropy-based reward, while Thompson sampling is used for arm selection. Thompson sampling provides a simple exploration-exploitation mechanism with negligible overhead \citep{Thompson1933ONTL,pmlr-v23-agrawal12}. The fallback arm is the \(K+1\)-th arm with \(\gamma_{K+1}=+\infty\). Operationally, it bypasses truncation and recovers the baseline sampler. This arm is important for stability in RL rollouts. Without it, training can begin with strong early gains but later overshoot in entropy, KL, gradient norm, log-ratio statistics, and related diagnostics because every rollout remains subject to some nonzero pruning pressure. The fallback arm gives the online controller a safe escape hatch when truncation is no longer beneficial.

\section{Method}

Let \(x=(x_1,\ldots,x_N)\) be a prompt and \(y_{<t}\) the generated sequence conditioned upon \(x\). At step \(t\), an LLM produces logits \(\ell_t \in \mathbb{R}^{|\mathcal{V}|}\). ANTS first computes a logit-space nucleus, \(\mathcal{N}_t\), masks logits outside the nucleus, and samples from the temperature-scaled distribution restricted to the surviving tokens:
\begin{equation}
\begin{gathered}
p_t^{\mathrm{ANTS}}(v) =
\frac{\mathbbm{1}\{v \in \mathcal{N}_t\}\exp(\ell_{t,v}/\tau)}
{\sum_{u \in \mathcal{N}_t}\exp(\ell_{t,u}/\tau)}.
\end{gathered}
\end{equation}
This design separates candidate-set selection from within-set stochasticity. That is, \(\mathcal{N}_t\) is determined by pre-temperature logits, while the scalar \(\tau\) still controls randomness among retained candidates.

\textbf{Entropy-conditioned logit-space nucleus.} For a selected arm \(a_t\), let \(\gamma(a_t)\) be its truncation parameter. This parameter defines the entropy-conditioned width \(n_t\) of the logit-space nucleus. Formally, we set
\begin{equation}
\begin{gathered}
n_t = n_0 + \gamma(a_t) \cdot\mathcal{H}(p_t^{(0)}), \qquad p_t^{(0)}=\texttt{softmax}(\ell_t),\quad \text{where }  \mathcal{H}(p)=-\sum_v p_v\log p_v.
\label{eq:entropy_threshold}
\end{gathered}
\end{equation}
The nucleus is then Equation~\ref{eq:ants_nucleus} with $n=n_t$. In practice $n_t$ is clipped to a valid range \([n_{\min}, n_{\max}]\) to prevent degenerate candidate sets. If the fallback arm is selected, we set \(\mathcal{N}_t=\mathcal{V}\), the vocabulary.

Entropy is reused as both a local width signal and a controller reward. The unit-temperature entropy \(\mathcal{H}(p_t^{(0)})\) determines the selected arm's token-level nucleus width in Equation~\ref{eq:entropy_threshold}, while the entropy of each arm-induced predictive distribution provides the intrinsic reward used for the bandit controller.

\textbf{Bandit controller.} We use a log-spaced grid of \(K>1\) truncation arms,
\begin{equation}
\gamma_k = 10^{\eta_k}, \qquad \eta_k = \log_{10}\gamma_{\min} + \frac{k -1 }{K - 1}\big(\log_{10}\gamma_{\max} - \log_{10}\gamma_{\min}\big),\qquad \forall\,k=1, \cdots, K,
\end{equation}
plus a fallback arm \(K+1\) with \(\gamma_{K+1}=+\infty\), and let \(\mathcal{A}=\{1,\cdots,K+1\}\) denote the set of all \(K+1\) arms, including the fallback arm. For bounded rewards \(r\in[0,1]\), each arm maintains a Beta posterior \((\alpha_k,\beta_k)\). At each sequence generation, the controller samples \(\tilde{q}_k \sim \texttt{Beta}(\alpha_k,\beta_k)\) and selects the arm \(a=\arg\max_k \tilde{q}_k\).
Because the nucleus sets for all arms can be constructed from the same sampled logits, we update every arm posterior at each step rather than observing feedback only for the selected arm. In principle, the reward for each arm can come from any bounded signal, including a task metric, verifier score, rule-based correctness signal, normalized judge score, or perplexity-based score. In our implementation, we use a lightweight intrinsic signal based on the entropy of the distribution induced by each arm. For each arm \(a\in\mathcal{A}\), we compute the entropy \(h_{a,t}\) over valid sampled decoding positions. We standardize these values within the current batch,
\begin{equation*}
\begin{gathered}
\mathcal{S}_h = \Big\{h_{b,s}, b\in\mathcal{A}\,\Big\vert\, m_{b,s}=1\Big\},\\
z_{a,t}=\frac{h_{a,t}-\mu_h}{\sigma_h+\epsilon},\qquad \mu_h=\texttt{mean}(\mathcal{S}_h),
\qquad
\sigma_h=\texttt{std}(\mathcal{S}_h).
\end{gathered}
\end{equation*}
Here \(m_{a,t}\) indicates whether arm \(a\) produced a valid nondegenerate nucleus at position \(t\). We then average over valid positions and map the averaged score to a bounded Thompson-sampling reward:
\[
r_a=\texttt{sigmoid}(\bar z_a), \qquad \bar z_a=\frac{\sum_t m_{a,t}z_{a,t}}{\sum_t m_{a,t}}.
\]
The posterior \((\alpha_a, \beta_a)\) of each arm \(a\in\mathcal{A}\) is updated as follows:
\[
\alpha_a \leftarrow 1 + (1-\rho)(\alpha_a-1) + r_a,
\qquad
\beta_a \leftarrow 1 + (1-\rho)(\beta_a-1) + (1-r_a),
\]
where \(\rho\) is a decay factor. Higher-than-average entropy therefore receives a larger posterior update, favoring arms that preserve sufficient support rather than arms that collapse the candidate set too aggressively. This reward is inexpensive because it is computed from the same arm-induced candidate distributions already constructed by the sampler, avoiding separate reward models, judges, or verifiers. When memory or inference overhead is not a constraint, however, the same update rule can instead use external task rewards or learned reward-model scores. The fallback posterior is updated in the same way as the finite truncation arms, allowing the controller to learn when no truncation is preferable.

\textbf{Fallback as stability control.} The fallback arm \(\gamma_{K+1}\) is more than a conservative option for inference. In rollout training, a sampler that is helpful early can become harmful later as the policy changes. If every arm applies finite truncation, the controller has no way to remove pruning pressure when rollout statistics begin to drift. Empirically, the no-fallback variant can start with strong gains but later overshoot in entropy, KL divergence, gradient norm, importance sampling or log ratio statistics, and related diagnostics. Treating \(\gamma_{K+1}=+\infty\) as a normal Thompson-sampling arm gives the online controller a learned escape hatch. Fundamentally, the controller can choose the baseline sampler when truncation is unsafe and return to finite truncation when the reward signal supports it.

\begin{algorithm}[t]
\caption{Adaptive Nucleus Truncation Sampling with Thompson Controller}
\label{alg:tins}
\begin{algorithmic}[1]
\State{Initialize \(K\) finite truncation arms \(\{\gamma_k\}_{k=1}^K\) and fallback arm \(K+1\) with \(\gamma_{K+1}=+\infty\).}
\State{Initialize Beta posteriors \((\alpha_k,\beta_k)=(1,1)\) for all arms.}
\For{each prompt \(x\)}
    \State{Sample \(\tilde{q}_k \sim \mathrm{Beta}(\alpha_k,\beta_k)\) and choose \(a=\arg\max_k\tilde{q}_k\).}
    \For{\(t=1,\ldots,T\)}
        \State{Compute logits \(\ell_t=f_\theta(x,y_{<t})\).}
        \If{\(a=K+1\)}
            \State{\(n_t \leftarrow +\infty\); \(\mathcal{N}_t \leftarrow \mathcal{V}\) \Comment{fallback/no truncation}.}
        \Else
            \State{\(n_t \leftarrow n_0 + \gamma(a)\cdot \mathcal{H}(\texttt{softmax}(\ell_t))\).}
            \State{\(\mathcal{N}_t \leftarrow \{v\in\mathcal{V}: \ell_{t,v} > \max_u \ell_{t,u} - n_t\sigma(\ell_t)\}\).}
        \EndIf
        \State{Sample \(y_t \sim \texttt{softmax}(\ell_t/\tau)\) restricted to \(\mathcal{N}_t\).}
    \EndFor
    % \State{Score the completed sequence and normalize reward \(r\in[0,1]\).}
    \State{Compute \(h_{a,t}, z_{a,t}, \bar{z}_{a,t}, r_a\) for all \(a\in\mathcal{A}\).}
    \State{\(\alpha_a\leftarrow \alpha_a + r_a\); \(\beta_a\leftarrow \beta_a + 1 - r_a,\,\forall\, a\in\mathcal{A}\).}
\EndFor
\end{algorithmic}
\end{algorithm}

\section{Experimental Setup}

\textbf{Models and regimes.} We evaluate ANTS on a 33B-total/4B-active sparse MoE reasoning architecture across 8K, 16K, and 32K generation budgets. For each budget, we compare ANTS against a baseline sampler that does not apply logit-space truncation. Curves are plotted over RL training checkpoints.

\textbf{Benchmarks.} We evaluate across tasks that expose different failure modes of long-budget generation: AIME 2024 and AIME 2025~\citep{aime24,aime25} for mathematical reasoning, GPQA Diamond~\citep{rein2023gpqa} for scientific QA, MMLU Pro~\citep{wang2024mmlu} for knowledge retrieval and multiple-choice reasoning, Codeforces ELO~\citep{quan2025codeelo} for executable code generation, IFBench Loose Prompt and Strict Prompt~\citep{pyatkin2025generalizing} for long-form instruction following, and IFEval Strict and Loose instruction scores for instruction compliance. AIME, GPQA, MMLU Pro, IFBench, and IFEval are reported in percentage points, while Codeforces is reported in ELO.

\textbf{Primary hypothesis.} Our guiding hypothesis is generation-budget scaling. Fundamentally, if ANTS suppresses noisy tails without removing useful reasoning paths, its benefit should increase as the generation budget grows and the cumulative risk of off-manifold continuation increases.

% \textbf{Implementation.} ANTS is applied to rollout generation during RL training rather than only at evaluation time. Throughout this stage, the Thompson sampling controller updates its arm beliefs in response to per-sequence task rewards and its state persists across the run. The baseline sampler is temperature-scaled sampling with no logit-space truncation. The bandit setting maintains \(K\) finite arms on a log-spaced \(\gamma\) grid in \([\gamma_{\min}, \gamma_{\max}]\), plus one fallback arm \(K+1\) with \(\gamma_{K+1}=+\infty\), with Beta posteriors initialized to \((\alpha, \beta) = (1, 1)\). The base threshold is \(n_0=0\) and, for finite arms only, the nucleus width is clipped to \([n_{\min}, n_{\max}]\) to prevent degenerate candidate sets. The fallback arm bypasses clipping and sets \(\mathcal{N}_t=\mathcal{V}\), the token vocabulary. Rewards are normalized to \([0, 1]\) using task verifier or rule-based correctness signals. Evaluations are logged every 80 training steps, and all curves and table entries reflect per-checkpoint averages over the full observed training window.

\textbf{Implementation.} ANTS is applied to rollout generation during RL training rather than only at evaluation time. The baseline sampler is temperature-scaled sampling with no logit-space truncation. The controller maintains \(K\) finite truncation arms on a log-spaced grid. Concretely, each arm stores a log-scale value \(\eta_k\in[\log_{10}\gamma_{\min},\log_{10}\gamma_{\max}]\), and the entropy multiplier used by the sampler is \(\gamma_k=10^{\eta_k}\). We set \(n_0=0, \gamma_{\min} = 5, \gamma_{\max}=32, K = 10\), and use the selected arm to form the entropy-conditioned width \(n_t=\gamma_k\mathcal{H}(p_t^{(0)})\). The finite-arm nucleus width is clipped to \([n_{\min},n_{\max}]\) to prevent degenerate candidate sets. We include a fallback arm \(K+1\), corresponding to \(\gamma_{K+1}=+\infty\), which bypasses clipping and sets \(\mathcal{N}_t=\mathcal{V}\), exactly recovering the no-truncation baseline over \(\mathcal{V}\).

The controller is updated using an intrinsic entropy-based reward rather than an external verifier or judge. For each arm, including the fallback arm, we compute the entropy of the arm-induced predictive distribution on sampled decoding positions. These entropy scores are standardized within the batch, averaged over valid positions, and passed through a sigmoid to obtain a bounded reward in \([0,1]\). Thus, arms with higher-than-average entropy receive larger posterior updates, favoring truncation levels that preserve sufficient support rather than collapsing the candidate set too aggressively. This reward is inexpensive because it reuses the candidate distributions already constructed for the different arms; no separate reward model, verifier, or judge is required. Evaluations are logged every 80 training steps, and all curves and table entries reflect per-checkpoint averages over the observed training window.

\section{Results}

We organize the evaluation around four questions. First, \emph{does ANTS improve as the generation budget grows?} Second, \emph{where does logit-space truncation have the most leverage?} Third, \emph{where is the sampler mostly irrelevant?} Finally, \emph{when can truncation be unsafe, and what does the fallback arm buy us during rollout training?} We provide answers to these questions in the following sections.
\begin{table}[t]
\centering
\small
\caption{ANTS improvement over the no-truncation baseline across generation budgets. Percentage metrics are absolute point differences. Codeforces is ELO difference. Positive deltas are bolded.}
\label{tab:main_deltas}
\begin{tabular}{lrrr}
\toprule
Benchmark & 8K & 16K & 32K \\
\midrule
AIME 2024 & \textbf{+4.2 pp} & \textbf{+2.2 pp} & \textbf{+3.1 pp} \\
AIME 2025 & \textbf{+3.7 pp} & \textbf{+4.0 pp} & \textbf{+7.0 pp} \\
GPQA Diamond & -0.0 pp & \textbf{+1.9 pp} & \textbf{+0.9 pp} \\
MMLU Pro & \textbf{+0.2 pp} & \textbf{+0.6 pp} & \textbf{+0.6 pp} \\
IFBench Loose Prompt & \textbf{+1.9 pp} & \textbf{+7.1 pp} & \textbf{+10.5 pp} \\
IFBench Strict Prompt & \textbf{+1.9 pp} & \textbf{+7.6 pp} & \textbf{+10.8 pp} \\
Codeforces & -59 ELO & \textbf{+230 ELO} & \textbf{+212 ELO} \\
IFEval Strict Inst. & \textbf{+0.3 pp} & \textbf{+0.6 pp} & \textbf{+1.5 pp} \\
IFEval Loose Inst. & \textbf{+0.1 pp} & \textbf{+0.6 pp} & \textbf{+1.2 pp} \\
\midrule
Mean (percentage metrics) & \textbf{+1.9 pp} & \textbf{+3.8 pp} & \textbf{+5.2 pp} \\
\bottomrule
\end{tabular}
\end{table}

\begin{figure}[t]
\centering
\includegraphics[width=0.98\linewidth]{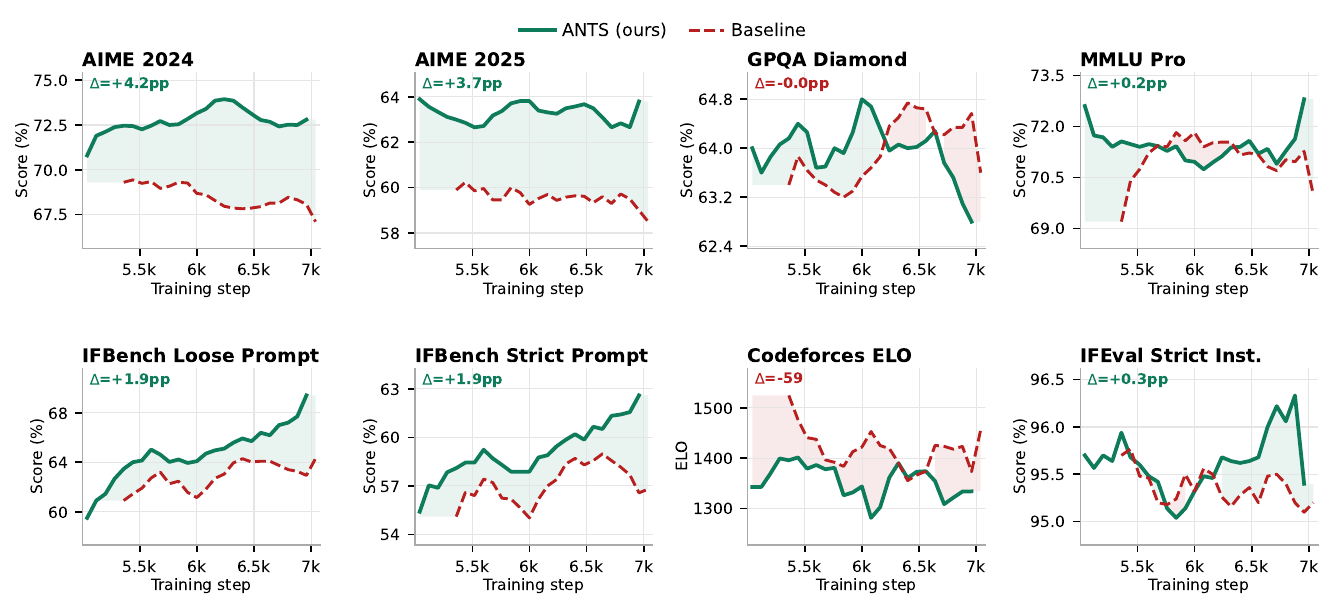}
\caption{8K generation budget on selected benchmarks. ANTS improves AIME and IFBench, is near parity on GPQA, and underperforms on Codeforces ELO. The Codeforces result indicates that ANTS is not uniformly beneficial under tight completion budgets.}
\label{fig:8k}
\end{figure}

\begin{figure}[t]
\centering
\includegraphics[width=0.98\linewidth]{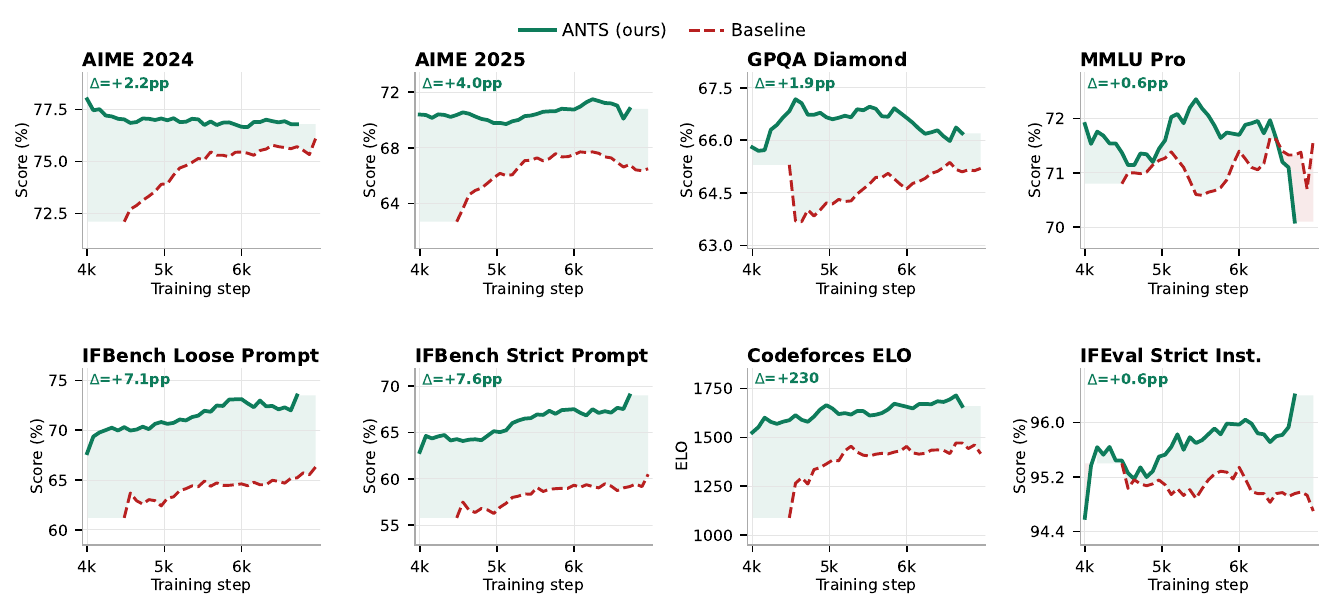}
\caption{16K generation budget on selected benchmarks. ANTS shows broad gains, including a large Codeforces reversal (+230 ELO) and strong IFBench improvements over the baseline.}
\label{fig:16k}
\end{figure}

\begin{figure}[t]
\centering
\includegraphics[width=0.98\linewidth]{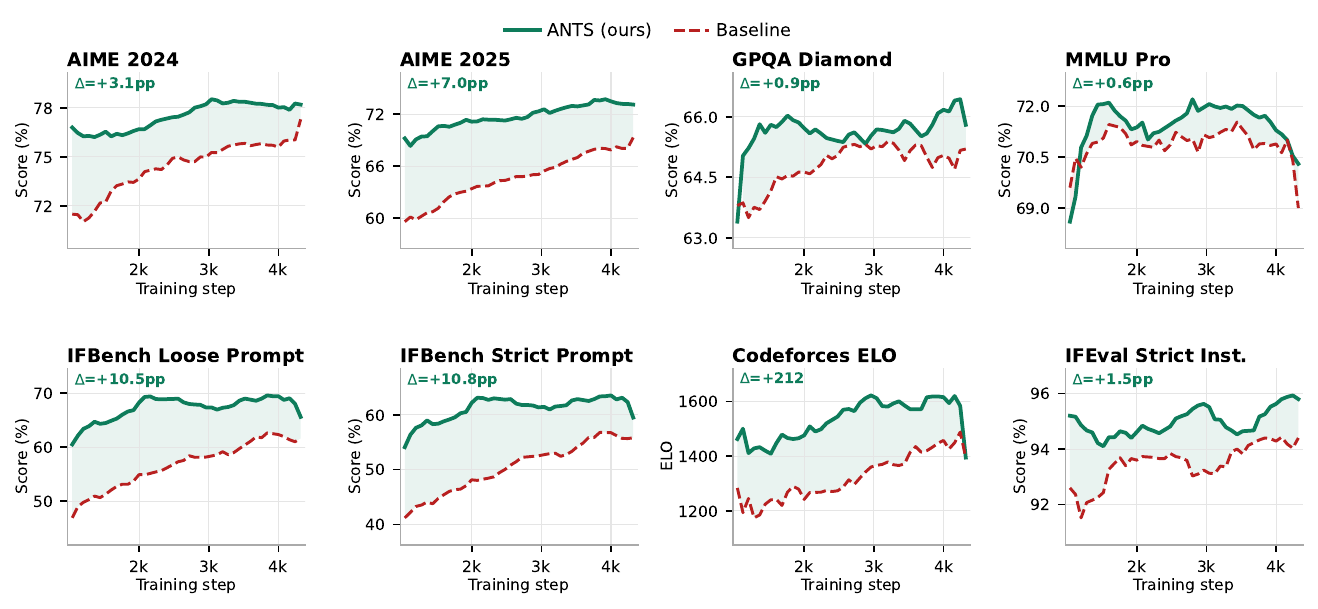}
\caption{32K generation budget on selected benchmarks. ANTS maintains a stable IFBench lead, improves AIME 2025 by 7.0 percentage points, and sustains a large Codeforces ELO gain.}
\label{fig:32k}
\end{figure}

\subsection{Generation-budget scaling}

The first question is whether the effect grows with budget rather than merely improving a few isolated benchmarks. Table~\ref{tab:main_deltas} shows that the mean gain over percentage-based benchmarks grows from \textbf{+1.9 points} at 8K to \textbf{+3.8} at 16K and \textbf{+5.2} at 32K. This trend holds after adding MMLU Pro and both IFEval instruction metrics, which are among the weakest results in the table and therefore provide a conservative test of the hypothesis. The strongest scaling appears on AIME 2025, which grows from \textbf{+3.7} to \textbf{+7.0 points}, and IFBench, which grows from \textbf{+1.9} to \textbf{+10.5}/\textbf{+10.8 points}. These are tasks where longer budgets may cause noisy, irrelevant, or overly verbose continuations to accumulate. The weakest scaling appears on MMLU Pro and IFEval, where gains are small because the tasks either reduce to short-form answer selection or are already near saturation. Both observations support the fact that ANTS helps most when the sampler has real leverage over a long generated trajectory.

\subsection{Mathematical reasoning}

AIME 2024 and AIME 2025 tell different stories. AIME 2024 peaks at 8K with \textbf{+4.2 points} and does not scale monotonically as it dips to \textbf{+2.2} at 16K before recovering to \textbf{+3.1} at 32K. AIME 2025 presents the cleaner scaling story. Gains grow from \textbf{+3.7} at 8K to \textbf{+4.0} at 16K, and \textbf{+7.0 points} at 32K, nearly doubling from 8K to 32K generation budgets. At 32K, ANTS continues climbing through much of training while the baseline occupies a lower band, suggesting that the average may understate the asymptotic gap if the baseline has not converged by the end of the observed window.

The asymmetry between the two AIME sets likely reflects differences in average solution length and problem distribution. If AIME 2025 requires longer reasoning chains on average, the per-step compounding benefit of suppressing unstable tails should grow with the available budget in a way that AIME 2024 does not fully expose. Disentangling solution length from year-to-year distributional shift would require per-problem difficulty and trace-length analysis, which we leave for future work.

\subsection{Instruction following}

Instruction following is where ANTS shows its strongest and most consistent gains. On the IFBench benchmark, Loose Prompt improves by \textbf{+1.9}, \textbf{+7.1}, and \textbf{+10.5 points} as the generation budget grows from 8K to 16K to 32K. Strict Prompt follows the same trajectory, improving by \textbf{+1.9}, \textbf{+7.6}, and \textbf{+10.8 points}. The 32K curves are particularly striking. ANTS holds a large lead from the earliest checkpoints, while the baseline improves gradually from a much lower starting point and does not close the gap within the observed window. A plausible explanation is that ANTS prevents the model from drifting into verbose or tangential continuations that violate exact instruction constraints. This failure mode becomes more likely as the generation budget increases, since longer responses provide more opportunities for noisy, unnecessary continuation, constraint drift, or off-task elaboration.

The IFEval benchmark provides a complementary view of the same phenomenon. Its Strict and Loose instruction scores are near saturation at 8K, with baseline scores ranging around 95--96\%, and therefore show correspondingly small gains: \textbf{+0.3} and \textbf{+0.1 points}. The 32K result is, however, more informative. ANTS improves Strict by \textbf{+1.5 points} and Loose by \textbf{+1.2 points}, and the improvement is accompanied by lower checkpoint-to-checkpoint variance. At 32K, the Strict score has a standard deviation of 0.91 under ANTS versus 1.30 under the baseline. The baseline occasionally dips below 92\%, while ANTS remains more steadily above 93\%. Thus, on IFEval, the value of ANTS is as much about stability as about mean accuracy. Put differently, logit-space truncation reduces tail events that cause occasional but sharp compliance failures when the model is given more room to generate.

\subsection{Code generation and budget-aware truncation}

\begin{figure}[t]
\centering
\includegraphics[width=0.98\linewidth]{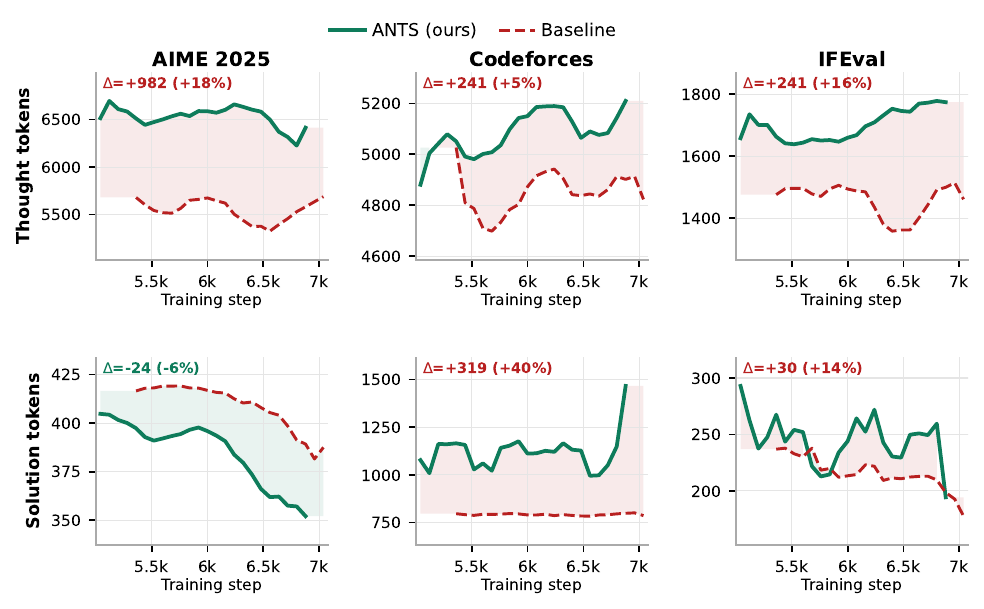}
\caption{Token allocation at the 8K budget. ANTS increases thought-token usage on AIME 2025, Codeforces, and IFEval, and produces substantially more Codeforces solution tokens despite lower 8K ELO. This shows that the 8K Codeforces gap is not simply a completion-length failure.}
\label{fig:tok_8k}
\end{figure}

Codeforces is the main caveat and the strongest evidence that truncation must be budget-aware. At 8K, ANTS underperforms by 59 ELO, the only benchmark-budget pair where it clearly falls below the baseline. At 16K and 32K, the result reverses sharply to \textbf{+230} and \textbf{+212 ELO}, respectively. Token analysis shows that the 8K failure is not caused by premature truncation of necessary code tokens. ANTS generates roughly 40\% more solution tokens at 8K while achieving lower ELO (see Figure~\ref{fig:tok_8k}). This result more likely points to a reasoning-quality problem rather than a completion-length problem.

The most likely explanation is that the 8K budget constrains the space of feasible complete programs, and ANTS reshapes the rollout distribution in a way that is less favorable in this constrained regime. At 16K and 32K completion lengths, complete programs become more reachable, and the diversity-quality benefit of suppressing unstable tails dominates throughout training. Codeforces therefore motivates the ``budget-aware'' part of ANTS, that is: \emph{truncation should be learned jointly with the available generation budget, task type, and training checkpoint rather than fixed globally.}

\begin{figure}[ht!]
\centering
\includegraphics[width=0.98\linewidth]{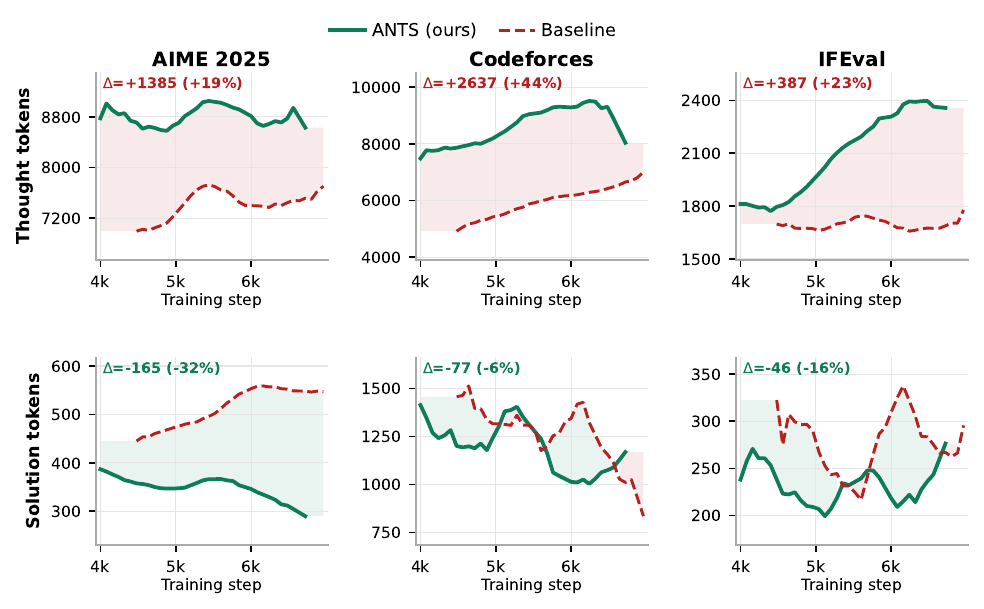}
\caption{Token allocation at 16K. ANTS increases thought token usage across all tasks, and more on Codeforces, while producing fewer solution tokens. This is consistent with a shift toward longer deliberation and more compact final answers in the regime where ELO reverses in favor of ANTS.}
\label{fig:tok_16k}
\end{figure}

\begin{figure}[ht!]
\centering
\includegraphics[width=0.98\linewidth]{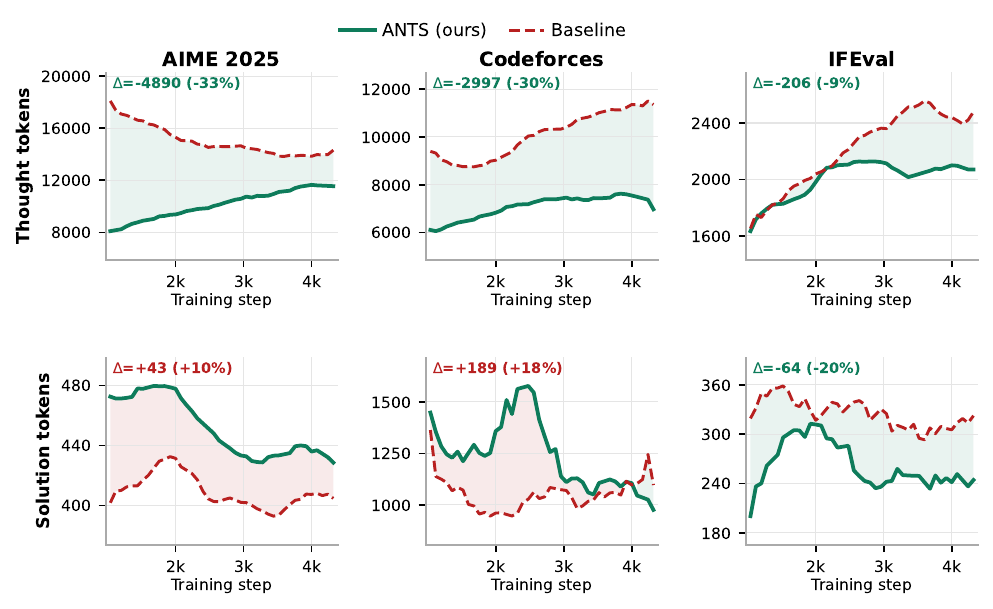}
\caption{Token allocation at 32K. ANTS uses fewer thought tokens across AIME 2025, Codeforces, and IFEval, while maintaining strong benchmark gains. This indicates that ANTS does not merely improve by generating longer reasoning traces. Its effect on token allocation is budget-dependent.}
\label{fig:tok_32k}
\end{figure}

\subsection{Token allocation and entropy-driven exploration}

Token statistics provide a useful view into how ANTS changes generation behavior beyond benchmark scores. Because the controller rewards higher-than-average entropy after standardization and sigmoid squashing, it favors arms that preserve sufficient support rather than arms that collapse the nucleus too aggressively. This does not directly optimize for longer outputs. Instead, it changes the distribution of reachable continuations, and the effect on token usage depends on the task and generation budget.

At 8K and 16K, ANTS generally increases thought token usage (see Figures~\ref{fig:tok_8k} and~\ref{fig:tok_16k}). On AIME 2025, thought tokens increase by \(18\%\) at 8K and \(19\%\) at 16K. On Codeforces, the increase is \(5\%\) at 8K and \(44\%\) at 16K. This is consistent with the controller preserving a broader exploratory support during intermediate reasoning. The solution-token pattern is more task-dependent. At 8K, ANTS produces \(40\%\) more Codeforces solution tokens while still trailing the baseline in ELO, which argues against the explanation that the 8K Codeforces gap is caused by premature shortening. At 16K generation budget, however, solution tokens decrease across AIME 2025, Codeforces, and IFEval, suggesting that ANTS can allocate more budget to deliberation while producing more compact final answers.

At 32K, the thought-token pattern reverses. ANTS uses fewer tokens than the baseline on AIME 2025, Codeforces, and IFEval, with reductions of \(33\%\), \(30\%\), and \(9\%\), respectively. Yet this is precisely the regime where ANTS achieves its largest average gains. This suggests that ANTS is not simply improving performance by generating longer chains of thought. Rather, its effect is budget-dependent. At moderate budgets, preserving entropy may encourage useful exploration. At long budgets, the same mechanism, together with the fallback arm, can suppress unnecessary continuation and stabilize the rollout distribution. The token results therefore support the broader interpretation that ANTS improves how the generation budget is allocated, not merely how much text the model produces.

\subsection{Scientific QA is near parity}

GPQA Diamond is near parity at 8K (\(-0.0\) points), improves at 16K (\textbf{+1.9 points}), and remains positive but smaller at 32K (\textbf{+0.9 points}). Thus, ANTS does help on GPQA, but the gain does not scale monotonically with generation budget. The 16K result suggests that logit-space truncation can improve scientific reasoning when the budget is large enough to support multi-step deliberation, while the smaller 32K gain indicates that this benefit may saturate or become more sensitive to checkpoint-level variation. We think that this is overall consistent with the benchmark format. GPQA tends to reward precise single-answer scientific reasoning, where the correct response often depends more on factual and conceptual precision than on avoiding long, open-ended continuations. As a result, reshaping the candidate set provides less leverage than it does on instruction following or long-form code generation, but remains directionally beneficial in the longer-budget regimes.

\subsection{Convergence behavior}

Beyond accuracy, ANTS often reaches its performance band earlier than the baseline. This is visible in AIME 2024 and IFBench at 16K and 32K, and most strikingly in 32K IFBench where ANTS occupies a higher band from the first recorded checkpoint onward. The baseline begins from a substantially lower level and closes toward ANTS gradually, but does not converge within the observed window.

Faster convergence is practically important in RL settings because rollout generation is expensive. Essentially, if a sampler reaches a target validation level with fewer training steps or fewer unstable trajectories, it can improve effective compute efficiency even when final accuracy gains are modest. The fallback arm is central to this stability story. Without fallback, training can initially improve but later overshoot in entropy, KL divergence, gradient norm, log-ratio statistics, and related diagnostics because every rollout remains subject to nonzero pruning pressure. The \(\gamma_{K+1}=+\infty\) arm gives the online controller a learned way to return to the untruncated sampler before these metrics run away.

This suggests that ANTS shapes the early-training rollout distribution in a way that is difficult for the baseline to recover from later. By suppressing unstable tails early, ANTS reduces the frequency of low-quality, irrelevant, or overly verbose rollouts that can destabilize sequence-level training signals. This effect is especially visible in the 32K IFBench curves, where ANTS starts in a higher performance band and the baseline improves only gradually. In this view, the practical benefit of ANTS is not only higher final accuracy, but also a more stable rollout distribution throughout training.

\section{Related Work}

\textbf{Truncation sampling.} Nucleus sampling was introduced to mitigate neural text degeneration by truncating unreliable probability tails while retaining diversity \citep{Holtzman2020The}. \cite{hewitt-etal-2022-truncation} framed truncation as desmoothing: the sampler should estimate the support of a cleaner true distribution rather than merely choose a probability prefix. Min-$p$ adapts the threshold using the top token's probability and reports gains in quality-diversity tradeoffs, especially at high temperature \citep{nguyen2025turningheatminpsampling}. Subsequent work has raised concerns about some empirical claims and the importance of comprehensive hyperparameter sweeps \citep{schaeffer2025minpmaxexaggerationcritical}. ANTS is complementary. It moves candidate-set construction from probability space to standardized logit space, then uses entropy conditioning, a bandit controller, and an explicit fallback arm to adapt strength. Relative to fixed top-\(n\sigma\), the main difference is that support size is no longer governed by one global \(n\); it can change with token-level uncertainty, task reward, checkpoint state, and budget.

\textbf{Logits as evidence.} Several works argue that logits preserve information that probability normalization can obscure~\citep{hinton2015distillingknowledgeneuralnetwork,guo2017calibrationmodernneuralnetworks}. \cite{tang-etal-2025-top} use logit-space separation to identify informative tokens for robust sampling. \cite{ma2025estimatingllmuncertaintyevidence} similarly argue that probability-based token uncertainty can lose evidence-strength information, motivating logit-based uncertainty estimation. ANTS uses this same high-level principle for rollout and evaluation-time sampling.

\textbf{Inference-time distribution manipulation.} Training-free or inference-time interventions modify model behavior without updating weights. SDA dynamically redistributes output probabilities according to alignment instructions without fine-tuning \citep{xia2026sda}. ANTS is complementary. Rather than steering generations toward explicit semantic attributes, it controls the reachable token set through standardized logit geometry, entropy conditioning, and a fallback arm for rollout stability.

\textbf{RL for reasoning.} Reasoning systems rely heavily on RL with rule-based rewards and long rollouts \citep{guo2025deepseek}. Dynamic rollout allocation and temperature scheduling have been proposed to improve RL efficiency and maintain exploration \citep{liao-etal-2025-enhancing}. ANTS provides a sampler-level mechanism for shaping rollout diversity without introducing additional policy-gradient terms.

\section{Limitations}

% The current evaluation is limited to one model family and a fixed set of tasks. The strongest conclusion is therefore not that ANTS universally dominates existing samplers, but that logit-space truncation exhibits a robust context-length scaling pattern on this model and benchmark suite. Codeforces at 8K shows that truncation can be harmful when the generation budget is too short. Future work should learn budget-conditioned fallback policies, evaluate calibration of arm rewards, and measure lexical reachability or coverage so that truncation does not silently remove rare but valid tokens. Recent work on word coverage suggests that standard filters can make plausible lexical choices unreachable, and similar audits would be valuable for ANTS \citep{awad2026lostsamplingassessinglexical}.

The current evaluation is limited to one MoE model family and a fixed set of tasks. The strongest conclusion is therefore not that ANTS universally dominates existing samplers, but that budget-aware logit-space truncation exhibits a robust generation-budget scaling pattern on this model and benchmark suite. Codeforces at 8K shows that truncation can be harmful in tightly budget-constrained code-generation regimes even when outputs are not shorter, so rollout length alone cannot explain all failures. Future work should learn budget-conditioned fallback policies, evaluate calibration of arm rewards, and measure lexical reachability or coverage so that truncation does not silently remove rare but valid tokens. Recent work on word coverage suggests that standard filters can make plausible lexical choices unreachable, and similar audits would be valuable for ANTS \citep{awad2026lostsamplingassessinglexical}.

\section{Conclusion}

We presented ANTS, a budget-aware logit-space sampler for long-form reasoning and RL rollouts. Unlike probability-space filters, ANTS selects its candidate set before temperature is applied. Unlike fixed top-\(n\sigma\), it does not rely on one global support width. Entropy conditioning adapts the nucleus at the token level, a Thompson controller adapts truncation strength across rollouts, and the \(\gamma_{K+1}=+\infty\) fallback arm allows the sampler to recover the untruncated baseline when finite truncation becomes unstable. On a 33B-total/4B-active MoE model, ANTS' average gains over percentage metrics increase from 1.9 points at 8K to 5.2 points at 32K, with the strongest effects on IFBench and AIME 2025, near-parity behavior on MMLU Pro, modest positive effects on GPQA, and a budget-dependent reversal on Codeforces. These results suggest that sampler design should be treated as part of inference-time scaling and rollout control, not merely as a fixed decoding hyperparameter.

% \clearpage
\Urlmuskip=0mu plus 1mu\relax
\bibliographystyle{plainnat}  %plain, unsrt, plainnat
\bibliography{refs}

\end{document}